# Automatic Facial Expression Recognition Using Features of Salient Facial Patches

S L Happy and Aurobinda Routray

**Abstract**— Extraction of discriminative features from salient facial patches plays a vital role in effective facial expression recognition. The accurate detection of facial landmarks improves the localization of the salient patches on face images. This paper proposes a novel framework for expression recognition by using appearance features of selected facial patches. A few prominent facial patches, depending on the position of facial landmarks, are extracted which are active during emotion elicitation. These active patches are further processed to obtain the salient patches which contain discriminative features for classification of each pair of expressions, thereby selecting different facial patches as salient for different pair of expression classes. One-against-one classification method is adopted using these features. In addition, an automated learning-free facial landmark detection technique has been proposed, which achieves similar performances as that of other state-of-art landmark detection methods, yet requires significantly less execution time. The proposed method is found to perform well consistently in different resolutions, hence, providing a solution for expression recognition in low resolution images. Experiments on CK+ and JAFFE facial expression databases show the effectiveness of the proposed system.

**Index Terms**—Facial expression analysis, facial landmark detection, feature selection, salient facial patches, low resolution image.

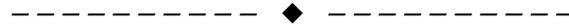

## 1 INTRODUCTION

Facial expression, being a fundamental mode of communicating human emotions, finds its applications in human-computer interaction (HCI), health-care, surveillance, driver safety, deceit detection etc. Tremendous success being achieved in the fields of face detection and face recognition, affective computing has received substantial attention among the researchers in the domain of computer vision. Signals, which can be used for affect recognition, include facial expression, paralinguistic features of speech, body language, physiological signals (e.g. Electromyogram (EMG), Electrocardiogram (ECG), Electrooculogram (EOG), Electroencephalography (EEG), Functional Magnetic Resonance Imaging (fMRI) etc.). A review of signals and methods for affective computing is reported in [1], according to which, most of the research on facial expression analysis are based on detection of basic emotions [2]: anger, fear, disgust, happiness, sadness, and surprise. A number of novel methodologies for facial expression recognition have been proposed over the last decade.

Effective expression analysis hugely depends upon the accurate representation of facial features. Facial Action Coding System (FACS) [3] represents face by measuring all visually observable facial movements in terms of Action Units (AUs) and associates them with the facial expressions. Accurate detection of AUs depends upon proper identification and tracking of different facial muscles irrespective of pose, face shape, illumination, and image resolution. According to Whitehill *et al.* [4], the detection of all facial fiducial points is even more challenging than expression recognition itself. Therefore, most of the existing algorithms are based on geometric and appearance based features. The models based on geometric features track the shape and size of the face and facial components such as eyes, lip corners, eyebrows etc., and categorize the expressions based on relative position of these facial components. Some researchers (e.g., [5], [6], [7], [8]) used shape models based on a set of characteristic points on the face to classify the expressions. However, these methods usually require very accurate and reliable detection as well as tracking of the facial landmarks which are difficult to achieve in many practical situations. Moreover, the distance between facial landmarks vary from person to person, thereby making the person independent expression recognition system less reliable. Facial expressions involve change in local texture. In appearance-based methods [9], a bank of filters such as Gabor wavelets, Local Binary Pattern (LBP) etc. are applied to either the whole-face or specific face regions to encode the texture. The superior performance of appearance based methods to the geometry based features is reported in [4]. The appearance-based methods generates high dimensional vector which are further represented in lower dimensional subspace by applying dimensionality reduction techniques, such as principal component analysis (PCA), linear discriminant analysis (LDA) etc. Finally, the classification is performed in learned subspace. Although the time and space costs are higher in appearance based methods, the preservation of discriminative information makes them very popular.

Extraction of facial features by dividing the face region into several blocks achieves better accuracy as reported by

---

- *S L Happy and A. Routray are with the Department of Electrical Engineering, Indian Institute of Technology, Kharagpur, India.*
  *E-mail: {happy,aroutray}@iitkgp.ac.in.*

many researchers ( [9], [10], [11], [12], [13], [14], [15]). However, this approach fails with improper face alignment and occlusions. Some earlier works [16], [17] on extraction of features from specific face regions mainly determine the facial regions which contributes more toward discrimination of expressions based on the training data. However, in these approaches, the positions and sizes of the facial patches vary according to the training data. Therefore, it is difficult to conceive a generic system using these approaches. In this paper, we propose a novel facial landmark detection technique as well as a salient patch based facial expression recognition framework with significant performance at different image resolutions. The proposed method localizes face as well as the facial landmark points in an image, thereby extracting some salient patches that are estimated during training stage. The appearance features from these patches are fed to a multi-class classifier to classify the images into six basic expression classes. It is found that the proposed facial landmark detection system performs similar to the state-of-the-art methods in near frontal images with lower computational complexity. The appearance features with lower number of histogram bins are used to reduce the computation. Empirically the salient facial patches are selected with predefined positions and sizes, which contribute significantly towards classification of one expression from others. Once the salient patches are selected, the expression recognition becomes easy irrespective of the data. Affective computing aims at effective emotion recognition in low resolution images. The experimental results shows that the proposed system performed better in low resolution images.

The paper is organized as follows. Section 2 presents a review of earlier works. The proposed framework is presented in Section 3. Section 4 and 5 discusses the facial landmark detection and feature extraction technique respectively. Experimental results and discussion are provided in Section 6. Section 7 concludes the paper.

## 2 RELATED WORK

For better performance in facial expression recognition, the importance of detection of facial landmarks is undeniable. Face alignment is an essential step and is usually carried out by detection and horizontal positioning of eyes. Facial landmark detection is followed by feature extraction. Selection of features also affects the classification accuracy. In [18], an active Infra-Red illumination along with Kalman filtering is used for accurate tracking facial components. Performance is improved by the use of both geometric and appearance features. Here the initial positions of facial landmarks are figured out using face geometry, given the position of eyes, which is not convenient. Tian *et al*. [19] also used relative distance (lip corner, eye, brow etc.) and transient features (wrinkles, furrows etc.) for recognizing AUs present in lower face. However, the use of Canny edge detector for extracting appearance features is not flexible in different illumination and determining the presence of furrows using threshold is uncertain. Uddin *et al*. [20] reported good performance by using image difference method for observing changes in expressions. The major issue is the landmark selection which is carried out manually by matching the eye and mouth regions. In [21], a relative geometrical distance based approach is described which uses computationally expensive Gabor filters for landmark detection and tracking. They used combined SVM and HMM models as classifiers.

Deformable models, to fit into new data instances, have become popular for facial landmark detection. Active shape models (ASM) determine shape, scale and pose by fitting an appropriate point distribution model (PDM) to the object of interest. Active appearance models (AAM) [22] combines both shape and texture models to represent the object, hence providing superior result to ASM. AAM is widely used ( [23], [24], [25], [26], [27]) for detection and tracking of non-rigid facial landmarks. However, its performance is poor in person independent scenarios. Manual placement of the landmark points in training data for construction of the shape model is a tedious task and time consuming process in these models. Constrained Local Model (CLM) framework proposed by Cristinacce *et al*. [28] has been proved as a better tool for person independent facial landmark detection. All the above said deformable models use PCA to learn the variability of shapes and textures offline. CLM algorithm is further modified by Saragih *et al*. [29] who proposed Regularized Landmark Mean Shift (RLMS) algorithm with improved landmark localization accuracy. Asthana *et al*. [30] proposed Discriminative Response Map Fitting (DRMF) method for the CLM framework for the generic face fitting scenario in both controlled and natural imaging conditions. Though satisfactory results has been achieved using these deformable models, high computational cost is an obstacle in using them in real-time applications. Chew *et al*. [31] established the fact that, appearance based models work robustly even with small alignment errors, and perform the same as that of a close to perfect alignment. Therefore, slight error in landmark detection will not hamper the purpose. In our experiments, we used a computationally inexpensive learning-free method for landmark detection that serves the purpose as efficiently as recent DRMF based CLM method [30].

An effective feature ideally discriminates between the expressions while minimizing the intra-class variance, and should be easily extracted from raw images of different resolutions. Among the appearance features, Gabor-wavelet representations have been widely adopted in face image analysis [32], [33] due to their superior performance. However, the computation of Gabor-features is both time and memory intensive; besides, they are sensitive to scaling. Recently the Local Binary Patterns (LBP) proved themselves as an effective appearance features for facial image analysis [10], [34], [35]. Jabid *et al*. [11] developed local facial descriptor based on Local Description Patterns (LDP) codes and obtained better performance than LBP features. Recently Dhall *et al*. [36] reported higher performance of Local Phase Quantization (LPQ) in facial expression recognition. In [12], Local Directional Pattern Variance (LDPv) is proposed which encodes contrast information using local variance of directional responses. However, Shan *et al*.

[37] found LBP features to be robust for analysis of low resolution images. Therefore, we used the LBP histograms as appearance features.

PCA ( [38], [39]) and LDA ( [40], [41], [42]) are used as a tool for dimensionality reduction as well as classification in expression recognition. In [43], authors reported the higher performance of PCA-LDA fusion method. An encrypted domain based facial expression recognition system is proposed in [44] which uses local fisher discriminant analysis to achieve accuracy as good as in normal images. Expression subspace is introduced in [45] which explains that the same expressions lie on the same subspace and new expressions can be generated from one image by projecting it into different emotion subspaces.

Most of the proposed methods use full face image, while a few use features extracted from specific facial patches. In [13], face image is divided into several sub regions (7x6) and local features (7x6x59 dimensional features) are extracted. Then, the discriminative LBP histogram bins are selected by using Adaboost technique for optimum classification. Similar approaches are reported in [14], [15], and [46]. In such cases, small misalignment would cause displacement of the sub region locations, thereby increasing error in classification. Moreover, for different persons the size and shape of facial organs are not the same, so, it cannot be assured that the same facial position always present in one particular block in all images. Hence, local patch selection based approach is adopted in our experiments. In [47], authors divided the face into 64 sub regions and explored the common facial patches which are active for most expressions and special facial patches which are active for specific expressions. Using multi task sparse learning method, they used features of a few number of facial patches to classify facial expressions. Song *et al.* [16] used eight facial patches based on specific landmark positions to observe the skin deformations caused by expressions. The authors have used binary classifiers to generate a Boolean variable for presence or absence of skin wrinkles. However, these patches do not include the texture of lip corners, which is important for expression recognition. Moreover, the occlusion of forehead by hair may result in false recognition. In [17], authors extracted Gabor features of different scales from the face image and trained using Adaboost to select the salient patches for each expression. However, the salient patch size and position is different when trained with different databases. Therefore, a unique criteria cannot be established for recognition of expressions in unknown images.

Some issues related to real-time detection of facial landmarks and expression recognition remain unaddressed so far. Most of the researches in this field are carried out on different datasets with suitable performance criteria befitting to the database. For example, selection of prominent facial areas improves the performance. However, in most of the literature, the size and position of these facial patches are reported to be different for different databases. Therefore, our experiments attempt to identify the salient facial areas having generalized discriminative features for expression classification. Selection of salient patches retaining discriminating features between each pair of facial expressions improved the accuracy. The size and location of patches are kept same for different databases for the purpose of generalization. In addition, the proposed framework has the potential to recognize expressions in low-resolution images.

## 3 Proposed Methodology

Changes in facial expressions involve contraction and expansion of facial muscles which alters the position of facial landmarks. Along with the facial muscles, the texture of the area also changes. This paper attempts to understand the contribution of different facial areas toward automatic expression recognition. In other words, the paper explores the facial patches which generates discriminative features to separate two expressions effectively.

The overview of the proposed method is shown in Fig. 1. Observations from [47], [16] suggest that accurate facial landmark detection and extraction of appearance features from active face regions improve the performance of expression recognition. Therefore, the first step is to localize the face followed by detection of the landmarks. A learning-free approach is proposed in which the eyes and nose are detected in the face image and a coarse region of interest (ROI) is marked around each. The lip and eyebrow corners are detected from respective ROIs. Locations of active patches are defined with respect to the location of landmarks. Fig. 2 shows the steps involved in automated facial landmark detection and active patch extraction. In training stage, all the active facial patches are evaluated and the ones having features of maximum variation between pairs of expressions are selected. These selected features are further projected into lower dimensional subspace and classified into different expressions using a multi-class classifier. The training phase includes pre-processing, selection of facial patches, extraction of appearance features and learning of the multi-class classifiers. In an unseen image, the process first detects the facial landmarks, then extracts the features from the selected salient patches, and finally classifies the expressions.

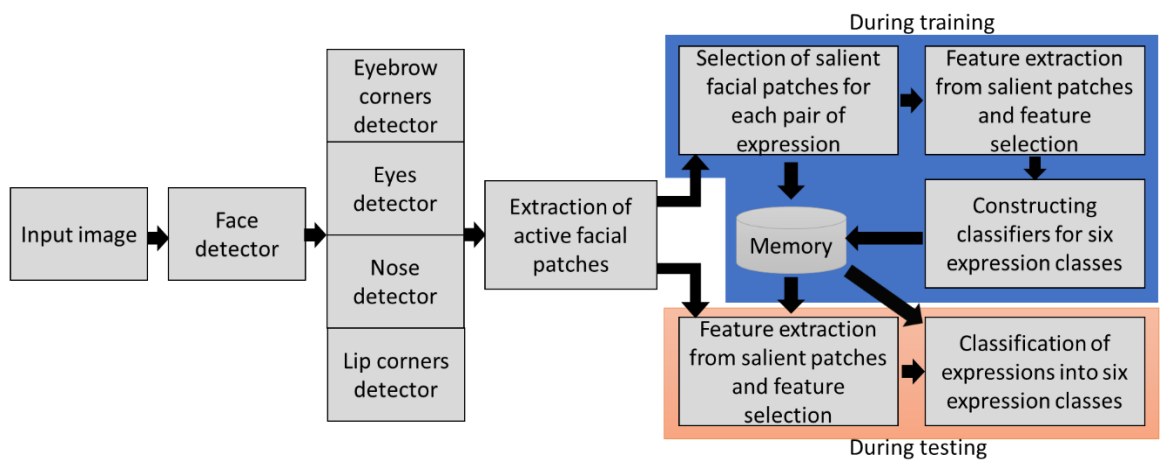

Fig. 1. Overview of the proposed system

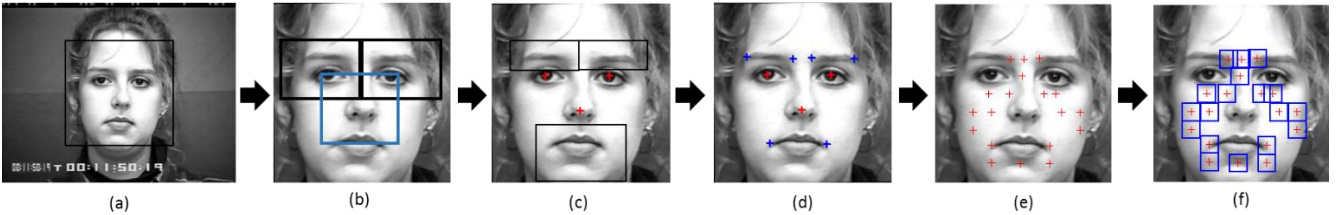

Fig. 2. Framework for automated facial landmark detection and active patch extraction, (a) face detection, (b) coarse ROI selection for eyes and nose, (c) eyes and nose detection followed by coarse ROI selection for eyebrows and lips, (d) detection of corners of lip and eyebrows, (e) finding the facial landmark locations, (f) extraction of active facial patches.

## 4 FACIAL LANDMARK DETECTION

The facial patches which are active during different facial expressions are studied in [47]. It is reported that some facial patches are common during elicitation of all basic expressions and some are confined to a single expression. The results indicate that these active patches are positioned below the eyes, in between the eyebrows, around the nose and mouth corners. To extract these patches from face image, we need to locate the facial components first followed by the extraction of the patches around these organs. Un-zueta et al. [48] proposed a robust, learning-free, lightweight generic face model fitting method for localization of the facial organs. Using local gradient analysis, this method finds the facial features and adjusts the deformable 3D face model so that its projection on image will match the facial feature points. In this paper, such a learning-free approach was adopted for localization of facial landmarks. We have extracted the active facial patches with respect to the position of eyes, eyebrows, nose, and lip corners using the geometrical statistics of the face.

### 4.1 Pre-processing

A low pass filtering was performed using a 3x3 Gaussian mask to remove noise from the facial images followed by face detection for face localization. We used Viola-Jones technique [49] of Haar-like features with Adaboost learning for face detection. It has lower computational complexity and was sufficiently accurate for detection of near-frontal and near-upright face images. Using integral image calculation, it can detect face regardless of scale and location in real time. The localized face was extracted and scaled to bring it to a common resolution. This made the algorithm shift invariant, i.e. insensitive to the location of the face on image. Histogram equalization was carried out for lighting corrections.

### 4.2 Eye and Nose Localization

To reduce the computational complexity as well as the false detection rate, the coarse region of interests (ROI) for eyes and nose were selected using geometrical positions of face. Both the eyes were detected separately using Haar classifiers trained for each eye. The Haar classifier returns the vertices of the rectangular area of detected eyes. The eye centers are computed as the mean of these coordinates. Similarly, nose position was also detected using Haar cascades. In our experiment, for more than 98% cases these parts were detected properly. In case the eyes or nose was not detected using Haar classifiers, the system relies on the landmark coordinates detected by anthropometric statistics of face. The position of eyes were used for up-right face alignment as the positions of eyes do not change with facial expressions.

### 4.3 Lip Corner Detection

Inspired by the work of Nguyen et al. [50], we used facial topographies for detection of lip and eyebrow corners. The ROIs for lips and eyebrows were selected as a function of face width positioned with respect to the facial organs. The ROI for mouth was extracted using the position of nose as reference (Fig. 3a). The upper lip always produces a distinct edge which can be detected using a horizontal edge detector. Sobel edge detector [51] was used for this purpose. In images with different expressions, a lot of edges were obtained which was further threshold by using Otsu method [52]. In this process, a binary image was obtained containing many connected regions. Using connected component analysis, the spurious components having an area less than a threshold were removed. Further, morphological dilation operation was carried out on the resulting binary image. Finally, the connected component with largest area which was just below the nose region was selected as upper lip region. Fig. 3 shows different stages of the process. The algorithm steps are given below.

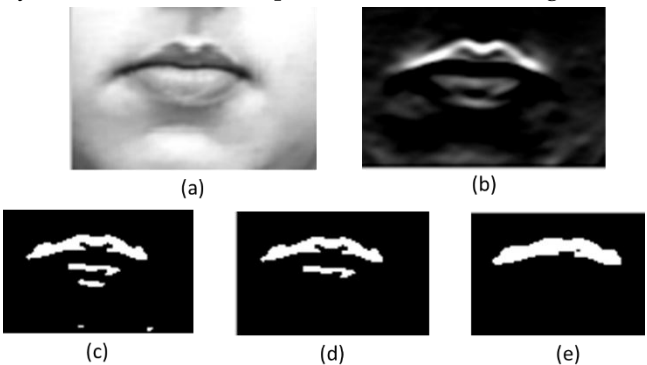

Fig. 3. Lip corner localization, (a) lips ROI, (b) applying horizontal Sobel edge detector, (c) applying Otsu threshold, (d) removing spurious, (e) applying morphological operations to render final connected component for lip corner localization.

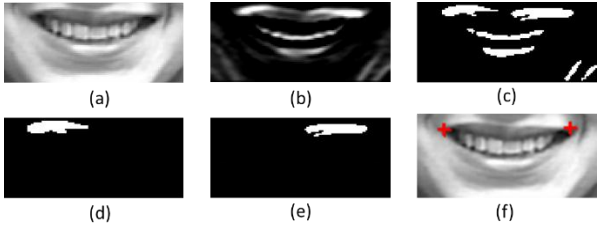

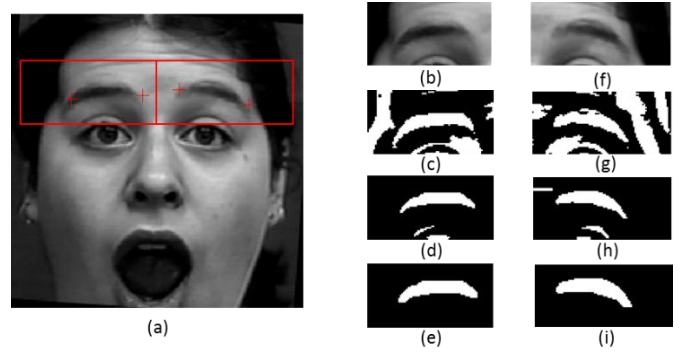

Fig. 4. Lip corner localization in a case where upper lip is not entirely connected, (a-c) same as Fig. 3, (d-e) selection of two connected components by scanning from top, (f) localized lip corner.

**Algorithm 1.** *Lip corner detection*

**Given:** aligned face ROI and nose position

1: select coarse lips ROI using face width and nose position
2: apply Gaussian blur to the lips ROI
3: apply horizontal sobel operator for edge detection
4: apply Otsu-thresholding
5: apply morphological dilation operation
6: find the connected components
7: remove the spurious connected components using threshold technique to the number of pixels
8: scan the image from the top and select the first connected component as upper lip position
9: locate the left and right most positions of connected component as lip corners

Sometimes, due to shadow below the nose, the upper lip could not be segmented properly. A case is shown in Fig. 4. In such cases, the upper lip was not segmented as a whole and the connected component obtained at the end resembled half of the upper lip. Hence, the extreme ends of this connected component did not satisfy the bilateral symmetry property, i.e. the lip corners should have been at more or less equal distances from vertical central line of face. These situations were detected by putting a threshold to the ratio of distance between the lip corners to the maximum of distances of the lip corners from the vertical central line. In such cases, the second connected component below the nose was considered as the other part of upper lip. Thus the lip corners were detected with the help of two connected components. By using the above said methods, false detection of lip corner points were minimized.

### 4.4 Eyebrow Corner Detection

With the knowledge of positions of eyes, the coarse ROIs of eyebrows were selected. The eyebrows were de-tected following the same steps as that of upper lip detection. However, we observed that performing an adaptive threshold operation before applying horizontal sobel operator improved the accuracy of eyebrow corner localization. The use of horizontal edge detector reduced the false detection of eyebrow positions due to partial occlusion by hair. The inner eyebrow corner was detected accurately in most of the images. Fig. 5 shows intermediate steps in eyebrow corner detection.

Fig. 5. Eyebrow corner localization, (a) rectangles showing search ROI and plus marks showing the detection result, (b & f) eye ROIs, (c & g) applying adaptive threshold on ROIs, (d & h) applying horizontal sobel edge detector followed by Otsu threshold and morphological operations, (e & i) final connected components for corner localization.

### 4.5 Extraction of Active Facial Patches

During an expression, the local patches were extracted from the face image depending upon the position of active facial muscles. We have considered the appearance of facial regions exhibiting considerable variations during one expression. For example, wrinkle in upper nose region is prominent in disgust expression and absent in other expressions. Similarly, regions around lip corners undergo significant changes and its appearance features are dissimilar for different expressions. From our observations, supported by the research of Zhong *et al*. [47], we used the active facial patches as shown in Fig. 6 for our experiment.

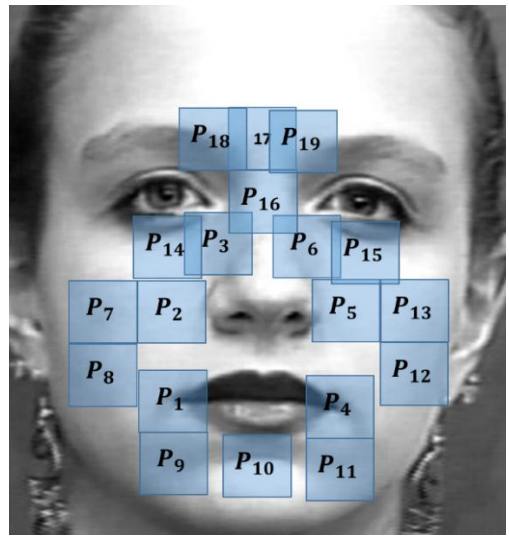

Fig. 6. Position of facial patches

The patches does not have very fixed position on the face image. Rather, their location depends upon the positions of facial landmarks. The size of all facial patches were kept equal and was approximately one-ninth of the width of the face. Here onwards, we will refer the patches by the numbers assigned to it. As shown in Fig. 6, $P_1$, $P_4$, $P_{18}$, and $P_{19}$ were directly extracted from the positions of lip corners and inner eyebrows respectively. $P_{16}$ was at the center of both the eyes; and $P_{17}$ was the patch above $P_{16}$. $P_3$ and $P_6$ were located in the midway of eye and nose. $P_{14}$ and $P_{15}$ were located just below eyes. $P_2$, $P_7$, and $P_8$ were clubbed together and located at one side of nose position. $P_9$ was located just below $P_1$. In a similar fashion $P_5$, $P_{11}$, $P_{12}$, and $P_{13}$ were located. $P_{10}$ was located at the center of position of $P_9$ and $P_{11}$.

## 5 FEATURE EXTRACTION AND CLASSIFICATION

LBP was widely used as a robust illumination invariant feature descriptor. This operator generates a binary number by comparing the neighbouring pixel values with the center pixel value [53]. The pattern with 8 neighborhoods is given by

$$LBP(x,y) = \sum_{n=0}^{7} s\,(i_n - i_c) 2^n$$

where $i_c$ is the pixel value at coordinate $(x, y)$ and $i_n$ are the pixel values at coordinates in the neighborhood of $(x, y)$, and

$$s(x) = \begin{cases} 1, x \geq 0 \\ 0, x < 0 \end{cases}$$

The histograms of LBP image can be utilized as feature descriptors, given by

$$H_i = \sum_{x,y} I\,\{\,LBP(x,y) = i\,\}, \qquad i = 0,1,\dots,n-1$$

where $n$ is the number of labels produced by LBP operator. Using different binwidths, the histograms can be grouped to discover different features. For instance, LBP with 8 neighboring points produces 256 labels. If we collect its histograms in 32 bins, then we are basically grouping the patterns [0,7], [8,15], [16,23], ..., and [248,255] together. This is same as ignoring the least significant bits of the unsigned integer, i.e. using patterns of one side of local neighborhood. Fig. 7 shows the pattern generated due to 32 histogram bins where the upper row neighbors do not contribute towards the pattern label. We used 16, 32 and 256 bin histograms in the experiments.

In addition, uniform LBP and rotation invariant uniform LBP values [53] are also used in our experiment and their performances are compared. Uniformity measure ($U$) corresponds to the number of bitwise transitions from 0 to 1 or vice-versa in a pattern when the bit pattern is traversed circularly. For instance, the pattern $(00000001)_2$ and $(00100110)_2$ have $U$ values 2 and 4 respectively. The pattern is called uniform (LBPu2) when $U \leq 2$. This reduces the length of the 8-neighborhood patterns to 59-bin histograms. The effect of rotation can be removed by assigning a unique identifier to each rotation invariant pattern, given by

$$LBPriu2 = \begin{cases} \sum_{n=0}^{7} s\,(i_n - i_c), if\ pattern\ is\ ''uniform'' \\ \qquad 9, \qquad otherwise \end{cases}$$

Thus, the rotational invariant uniform LBP with 8 neighborhood produces 10 histogram bins.

### 5.1 Learning Salient Facial Patches Across Expressions

In most of the literatures, all the facial features are concatenated to recognize the expression. However, this generates a feature vector of high dimension. We observed that the features from a fewer facial patches can replace the high dimensional features without significant diminution of the recognition accuracy. From human perception, not all facial patches are responsible for recognition of one expression. The facial patches responsible for recognition of each expression can be used separately to recognize that particular expression. Based on this hypothesis, we evaluated the performance of each facial patch for recognition of different expressions.

Further, some expressions share similar movements of facial muscles; features of such patches are redundant while classifying the expressions. Therefore, after extracting the active facial patches, we selected the salient facial patches responsible for discrimination between each pair of basic expressions. A facial patch is considered to be discriminative between two expressions, if the features extracted from this patch can classify the two expressions accurately. Note that not all active patches are salient for recognition of all expressions. For all possible pair of expressions ($_6C_2$ ), all the 19 active patches were evaluated by conducting a ten-fold cross validation test. The patches that result maximum discrimination were selected for representing the expressions.

The LBP histogram features in lower resolution images are sparse in nature because of the smaller patch area. LDA was applied for projecting these features to the discriminating dimensions and to choose the salient patches according to their discriminative performance. LDA finds the hyper-plane that minimizes the intra-class scatter ($S_w$), while maximizing the inter-class scatter ($S_b$). It is also used as a tool for interpretation of importance of the features. Hence it can be considered as a transformation into a lower dimensional space for optimal discrimination between classes. The intra-class scatter ($S_w$) and inter-class scatter ($S_b$) are given by

$$S_b = \sum_{i=1}^{n} N_i\,(\bar{x}_i - \bar{x})(\bar{x}_i - \bar{x})^T$$

$$S_w = \frac{1}{N_i - 1} \sum_{i=1}^{n} \sum_{j=1}^{N_i} (x_{i,j} - \bar{x}_i)(x_{i,j} - \bar{x}_i)^T$$

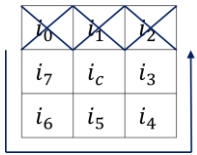

Fig. 7. Patterns generated by one side of local neighborhood which produces 32 histogram bins

where $x_{i,j}$ is the $j$th feature vector in $i$th class. Here $n$ number of classes, having $N_i$ number of images in $i$th class, have $\bar{x}$, the mean vector of all the training data, and $\bar{x}_i$, the mean of $i$th class. LDA aims at maximization of $S_b$ while minimizing $S_w$, i.e. maximization of ratio of determinant of $S_b$ to the determinant of $S_w$.

$$\Phi_{lda} = \frac{\arg max}{\Phi} \frac{|\Phi^T S_b \Phi|}{|\Phi^T S_w \Phi|}$$

This ratio is called as Fishers criterion. This can be computed by solving the generalized eigenvalue problem given as:

$$S_b \Phi_{lda} - S_w \Phi_{lda} \Lambda = 0$$
$$\Rightarrow S_w^{-1} S_b \Phi_{lda} = \Phi_{lda} \Lambda$$

where $\Lambda$ is the diagonal eigenvalue matrix and $\Phi_{lda}$ is the set of discriminant vectors of $S_b$ and $S_w$ corresponding to the $n-1$ largest generalized eigenvalues. Thus, Fisher criterion is maximized when the projection matrix $\Phi_{lda}$ is composed of eigenvectors of $S_w^{-1}S_b$, subject to $S_w$ being non-singular. As suggested by Belhumeur *et al.* [54], PCA was applied to the signal prior to LDA. By doing so, the signal was projected to lower dimensional space assuring the non-singularity of within class scatter matrix. Moreover, this PCA-LDA fusion [43] improves the performance. Therefore, we applied PCA on the training set for dimensionality reduction followed by LDA.

We calculated the saliency of all facial patches for all pair of expressions and it was expressed in terms of saliency scores. The saliency of a patch represents the ability of the features from the patch to accurately classify a pair of expressions. The saliency score of a patch between a pair of expressions is the classification accuracy of the features from that patch in classifying the two expressions. Here PCA-LDA was used for classification purpose to determine the saliency score. In a similar fashion, saliency score of all patches for each pair of expressions were calculated. We used one-against-one strategy for expression classification purpose. While classifying between a pair of expressions, the features extracted from those facial patches which have high saliency score. The feature vectors from the salient patches were concatenated to construct a higher dimensional feature vector. Thus, the dimension of the feature vector depends upon the number of patches selected for classification purpose.

We applied PCA to reduce the dimensionality of the feature vector. Thus, by projecting the feature vectors from salient patches to the optimal sub-space obtained by above method, we can find the lower dimensional vector with maximum discrimination for different classes. The weight vectors, corresponding to the salient patches of each pair of expression classes, generated during the training stage were used during testing.

## 5.2 Multi-class Classification

SVM was used for classification of extracted features into different expression categories. SVM [55] is a popular machine learning algorithm which maps the feature vector to a different plane, usually to a higher dimensional plane, by a non-linear mapping, and finds a linear decision hyper plane for classification of two classes. Since SVM is a binary classifier, we implemented one-against-one (OAO) technique for multi-class classification [56]. In OAO approach, a classifier is trained between each pair of classes; hence $_KC_2$ number of classifiers were constructed in total, where $K$ is the number of classes. Using voting strategy, a vector can be classified to the class having the highest number of votes. After several experiments with linear, polynomial, and radial basis function (RBF) kernels, we selected RBF kernels for its superior classification performance.

## 6 Experiments And Discussion

The proposed method was evaluated by using two widely used facial expression databases, i.e., Japanese Female Facial Expressions (JAFFE) [57] and Cohn-Kanade (CK+) [58]. We have used ten-fold cross validation to evaluate the performance of the proposed method. As discussed earlier, face detection was carried out on all images followed by scaling to bring the face to a common resolution. Facial landmarks were detected and salient facial patches were extracted from each face image. During training stage, a SVM classifier was trained between each pair of expressions. Here the training data were the concatenated LBP histogram features extracted from the salient patches containing discriminative characteristics between the given pair of expression classes. Similarly, $_6C_2$ numbers of SVM classifiers were constructed and used for evaluating the performance on the test-set.

### 6.1 Experiments on the Cohn-Kanade Database

The Cohn-Kanade database contains both male and female facial expression image sequences for the six basic emotions. In our experiments, the last image from each sequence was selected where the expression is at its peak intensity. The number of instances for each expression varies according to its availability. In our experiments on CK+ database, we used 329 images in total: anger (41), disgust (45), fear (53), happiness (69), sadness (56), and surprise (65).

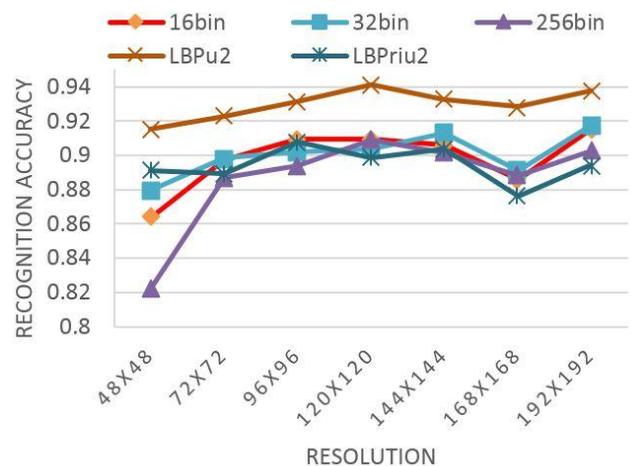

Fig. 8. The recognition rate in images with different face resolutions in CK+ database

### 6.1.1 Analysis of histogram binwidth and face resolution

We determined the optimal resolution and binwidth of the histogram empirically. The expression recognition performance of different feature vectors were studied for face resolutions starting from 48x48 to 192x192. The low resolution images were obtained by down sampling the images. The classifiers were trained and evaluated at different resolutions of face images. In our experiments, we observed minimum accuracy of 82% at 48x48 face resolution as shown in Fig. 8. In [59], it is reported that with a face image of 48x64 resolution, some facial landmarks, such as, lip and eye corners are difficult to detect. Therefore, it is uncertain if expressions can be recognized at this resolution. However, from our experiment, we obtained pretty good accuracy at all resolutions. This establishes the robustness of the appearance features extracted from the salient patches at different resolutions. Since we have implemented a voting method, the result is based on votes of $_6C_2$ classifiers, threby, reducing the classification error due to a single classifier.

We also observed that the use of LBPu2 features produced better accuracy compared to other features at all resolutions. The performance of the other feature vectors are alike. The uniform patterns removes the noisy estimates in the image by accumulating them into one histogram bin, thereby increasing the recognition accuracy.

### 6.1.2 Performance improvement: use of block-histograms

By implementing block-based feature extraction technique, more local features were added to the feature vector. This process makes the feature vector to be a combination of local as well as global features. Empirically we observed that the performance was improved when each selected patch was further divided into four equal blocks. In our experiments, the feature vector was obtained by concatenating the features obtained from each block of the salient patches and the results are shown in Fig. 9. It was observed that the 16-bin histograms, 32-bin histograms, 256-bin histograms, and uniform LBP features performed alike at all resolutions. We performed the experiments on a standard face resolution of 96x96. At this resolution, all the features except LBPriu2 had similar performances. Feature vector with low dimension reduces the computational complexity. Therefore, we selected the histogram features with 16-bins as the optimal trade-off between speed and accuracy. Table 1 shows the confusion matrix of six emotions based on the proposed method. The quality of the overall classification is evaluated by calculating the macro-average [60] of precision, recall and F-score. The proposed system attained a balanced F-score of 94.39% with 94.1% recall and 94.69% precision.

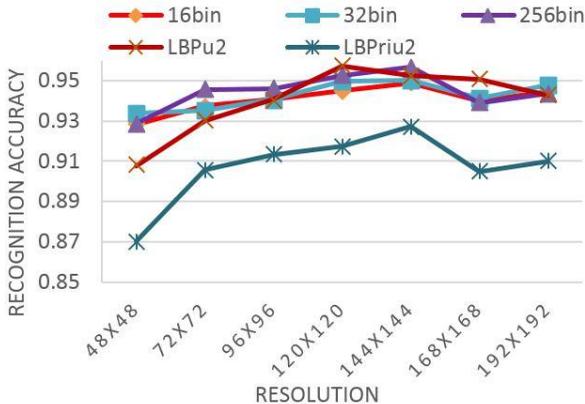

Fig. 9. Improvement in the recognition rate at different resolutions by using block histograms

|  | Anger | Fear | Disgust | Happiness | Sadness | Surprise |
|---|---|---|---|---|---|---|
| **Anger** | **87.8** | 0 | 0 | 0 | 7.32 | 4.88 |
| **Fear** | 0 | **93.33** | 0 | 4.44 | 0 | 2.22 |
| **Disgust** | 0 | 1.88 | **94.33** | 0 | 1.88 | 1.88 |
| **Happiness** | 1.44 | 2.89 | 0 | **94.2** | 0 | 1.44 |
| **Sadness** | 1.78 | 0 | 0 | 1.78 | **96.42** | 0 |
| **Surprise** | 0 | 0 | 0 | 1.53 | 0 | **98.46** |

As observed from Table 1, surprise expression achieved best recognition rate which is usually characterized by open mouth and upward eyebrow movement. The system performed worst for anger expression and classification error was maximum between anger and sadness since they involve similar and subtle changes. Here onwards, all the experiments are based on face resolution of 96x96.

### 6.1.3 Optimum number of salient patches

Number of patches used for classification also affects the performance in terms of speed and accuracy. Fig. 10 shows average of accuracies of all expressions with respect to the number of salient patches used for classification with a face resolution of 96x96. It is apparent from Fig. 10 that the use of features from all the 19 patches can classify all expressions with an accuracy of 93.87%. It is clear that even the use of appearance features of a single salient patch can discriminate between each pair expressions efficiently with recognition rate of 91.19%. This implies that the use of rest of the features from other patches contribute minimum towards the discriminative features. More the number of patches used, more is the size of the feature vector.

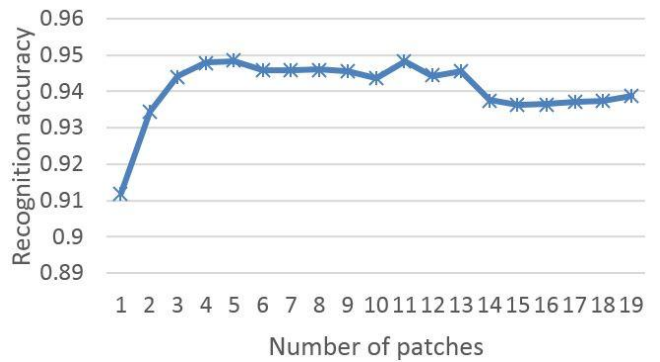

Fig. 10. The recognition rate using different number of salient patches

This increases the computational burden. Therefore, instead of using all the facial patches, we can rely on some salient facial patches for expression recognition. This will improve the computational complexity as well as robustness of the features especially when a face is partially occluded. In our experiments, we used top four salient patches in our experiments which results an accuracy close to 95%. Note that the combination of patches varies across different expression pairs.

### 6.1.4 Performance comparison

The proposed method was compared with the results obtained by other approaches reported in the literature. Lack of the knowledge of the data and evaluation protocol used by different literatures makes the comparison task difficult. However, we compared the performance of the system with literatures that adopted similar protocols in CK+ dataset. Table 2 compares the performance of the proposed method with the state-of-the-art methods. From Table 2, Uddin *et al.* [20] reported highest recognition performance for disgust, fear, and happiness probably due to the use of temporal features through Hidden Markov Model. However, the performance of the proposed system is comparable with the other systems as it achieved an average recognition rate of 94.09%. Nevertheless, the high recognition rate is obtained using features from specific facial patches and without using features of temporal domain.

## 6.2 Experiments on JAFFE Database

While testing on JAFFE database, we used the same pa-rameters obtained for Cohn-Kanade database. In our experiments on JAFFE database, we used 183 images in total: anger (30), disgust (32), fear (29), happiness (31), sadness (31), and surprise (30). The confusion matrix, as in Table 3, shows the consistent performance of the proposed method. An overall accuracy of 91.8% was obtained. From the experiments on JAFFE database, it was observed that the proposed system recognises all expressions with 91.8% recall and 92.63% precision achieving an average F-score of 92.22%. The system performed worst for sadness expression as it misclassified sadness as anger.

## 6.3 Experiments on Fused Database

For generalization, we have fused the samples of two databases together to train the classifier [62]. Sample Level fusion was performed by putting the images of both databases together. The training set was constructed by randomly electing 90% of the data from each expression of each database. The rest data were used as testing set. The models were trained, and their performances were evaluated on samples of individual databases in testing set. This experiment was repeated for ten times. By learning the features from different databases, the classifier performs better in various situations. All samples were treated with

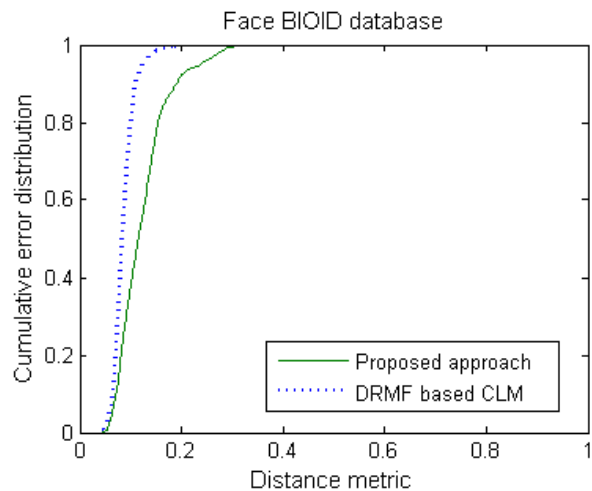

Fig. 11. Comparison between performance of the proposed landmark detection method and the DRMF based CLM model in BIOID database.

#### TABLE 2
PERFORMANCE COMPARISON OF DIFFERENT STATE-OF-THE-ART APPROACHES ON CK+ DATABASE

|  | [20] | [61] | [47] | [16] | [17] | Proposed system |
|---|---|---|---|---|---|---|
| **An** | 82.5 | 87.03 | 71.39 | **90.56** | 87.1 | 87.8 |
| **Di** | **97.5** | 91.58 | 95.33 | 86.04 | 90.2 | 93.33 |
| **Fe** | **95** | 90.98 | 81.11 | 84.61 | 92 | 94.33 |
| **Ha** | **100** | 96.92 | 95.42 | 93.61 | 98.07 | 94.2 |
| **Sa** | 92.5 | 84.58 | 88.01 | 90.24 | 91.47 | **96.42** |
| **Su** | 92.5 | 91.23 | 98.27 | 92.3 | **100** | 98.46 |
| **Avg** | 93.33 | 90.38 | 88.255 | 89.56 | 93.14 | **94.09** |

An = anger, Di = disgust, Fe = fear, Ha = happiness, Sa = sadness, Su = surprise, and Avg = Average recognition rate.

#### TABLE 3
THE CONFUSION MATRIX USING PROPOSED METHOD ON *JAFFE* DATABASE

|  | Anger | Fear | Disgust | Happiness | Sadness | Surprise |
|---|---|---|---|---|---|---|
| **Anger** | 100 | 0 | 0 | 0 | 0 | 0 |
| **Fear** | 0 | **93.75** | 0 | 0 | 0 | 6.25 |
| **Disgust** | 6.89 | 6.89 | **86.2** | 0 | 0 | 0 |
| **Happiness** | 0 | 0 | 0 | **96.77** | 0 | 3.22 |
| **Sadness** | 9.67 | 6.45 | 0 | 6.45 | **77.41** | 0 |
| **Surprise** | 0 | 3.33 | 0 | 0 | 0 | **96.66** |

#### TABLE 4
COMPARISON BETWEEN PERFORMANCE AND TIME COMPLEXITY OF THE PROPOSED LANDMARK DETECTION METHOD AND THE DRMF BASED CLM MODEL ON CK+ DATABASE

| Landmark detection method | CLM using DRMF | Our method |
|---|---|---|
| Expression Recognition accuracy (%) | 92.42 | 94.14 |
| Time(in sec. per image) | 1.6746 | 0.2955 |

(Recognition accuracy is obtained by using the proposed salient patch extraction based method. The execution time analysis is based on unoptimized MATLAB code in a Dual-Core Intel Pentium i5 CPU with 3.2 GHz.)

TABLE 5
THE SALIENT PATCHES DERIVED FROM FUSION OF CK+ AND JAFFE DATABASES

| | Anger | Fear | Disgust | Happiness | Sadness | Surprise |
|---|---|---|---|---|---|---|
| **Anger** | – | $P_1, P_4, P_9, P_{10}$ | $P_2, P_4, P_5, P_6$ | $P_1, P_4, P_9, P_{11}$ | $P_1, P_9, P_{10}, P_{18}$ | $P_1, P_4, P_9, P_{10}$ |
| **Fear** | – | – | $P_1, P_2, P_4, P_8$ | $P_1, P_4, P_8, P_9$ | $P_1, P_4, P_8, P_9$ | $P_1, P_5, P_{11}, P_{12}$ |
| **Disgust** | – | – | – | $P_1, P_4, P_5, P_6$ | $P_1, P_9, P_{18}, P_2$ | $P_1, P_2, P_5, P_6$ |
| **Happiness** | – | – | – | – | $P_1, P_7, P_9, P_{11}$ | $P_2, P_4, P_5, P_{11}$ |
| **Sadness** | – | – | – | – | – | $P_1, P_9, P_{10}, P_{11}$ |
| **Surprise** | – | – | – | – | – | – |

equal probability of selection for training or testing. Therefore, it is expected that the database with more samples should dominate in performance. However, the proposed method performed well on both databases with a significant accuracy of 89.64% and 85.06% on CK+ and JAFFE databases respectively. The top four salient patches for classification of each pair of expressions is provided in Table 5.

## 6.4 Performance of the landmark detector

We have used the images of BIOID [63] dataset without spectacles for evaluating the performance of the proposed landmark detection algorithm. It contains manually labelled facial landmarks which serves the purpose of ground truth during training and testing. The average Euclidian distance error from point to point for each landmark location is used as the distance measure, which is given as:

$$e = \frac{1}{ns} \sum_{i=1}^{n} d_i$$

Here $d_i$s are the Euclidean distance errors for the landmarks, $n$ is the number of landmarks, and $s$ is the distance between the eyes pupils used as the scaling factor. Fig. 11 shows the cumulative distribution of the detection accuracy by using proposed method. Its performance was compared with the performance of the recent DRMF method based CLM model [30]. The performance of expression recognition by the two landmark detection methods was also compared. After detection of facial landmarks, the active patches were extracted and the procedures as discussed in section 5 were followed for expression classification. As shown in Table 4, both the methods produce almost similar accuracy in CK+ database while the proposed landmark detection method takes very small time in comparison to the other one. Although the DRMF based CLM method finds the facial organ locations accurately, it sometimes fails to fit to their shapes in different expressions, especially on lips. When the mouth is open or tightly closed, or when the lip corners are pulled down, it at times generates average lip-shape instead of completely fitting to lips. Thus, lip corner patches were extracted at incorrect locations resulting poor classification. However, the proposed landmark detection method accurately finds the facial landmarks in most of the images thereby extracting features from appropriate patch locations. The slight misalignment error is automatically taken care by the appearance feature.

## 7 CONCLUSION

This paper has presented a computationally efficient facial expression recognition system for accurate classification of the six universal expressions. It investigates the relevance of different facial patches in the recognition of different facial expressions. All major active regions on face are extracted which are responsible for the face deformation during an expression. The position and size of these active regions are predefined. The system analyses the active patches and determines the salient areas on face where the features are discriminative for different expressions. Using the appearance features from the salient patches, the system performs the one-against-one classification task and determines the expression based on majority vote.

In addition, a facial landmark detection method is described which detects some facial points accurately with less computational cost. Expression recognition is carried out using the proposed landmark detection method as well as the recently proposed CLM model based on DRMF method. In both cases, recognition accuracy is almost similar, whereas computational cost of the proposed learning-free method is significantly less. Promising results has been obtained by using block based LBP histogram features of the salient patches. Extensive experiments has been carried out on two facial expression databases and the combined dataset. Experiments are conducted using various binwidths of LBP histograms, uniform LBP and rotation invariant LBP features. Low dimensional features are preferred, with sufficient recognition accuracy, to decrease computational complexity. Therefore, the 16-bin LBP histogram features are used from four salient patches at face resolution of 96x96 for obtaining best performance with a suitable trade-off between speed and accuracy. Our system found the classification between anger and sadness troublesome in all databases. The system appears to perform well in CK+ dataset with an F-score of 94.39%. Using the salient patches obtained by training on CK+ dataset, the system achieves an F-score of 92.22% in JAFFE dataset. This proves the generic performance of the system. The performance of the proposed system is comparable with the earlier works with similar approach, nevertheless our system is fully automated.

Interestingly, the local features at the salient patches provide consistent performance at different resolutions. Thus, the proposed method can be employed in real-world applications with low resolution imaging in real-time. Security cameras, for instance, provide low resolution images which can be analysed effectively using the proposed framework.

Instead of the whole face, the proposed method classifies the emotion by assessing a few facial patches. Thus, there is a chance of improvement in performance with partially occluded images which has not been addressed in this study. This analysis is confined to databases without facial hairs. There is a possibility of improvement by using different appearance features. Dynamics of expression in temporal domain is also not considered in this study. It would be interesting to explore the system incorporated with motion features from different facial patches. The execution time reported for the proposed algorithm is based on an un-optimized MATLAB code. However, the optimal implementation of the proposed framework will significantly improve the computational cost and real-time expression recognition can be achieved with substantial accuracy. Further analysis and efforts are required to improve the performance by addressing some of the above mentioned issues.

## 8 ACKNOWLEDGMENT

The authors would like to thank Prof. Jeffery Cohn for the use of the Cohn–Kanade database, and Dr. Michael J. Lyons for the use of the JAFFE database. The authors gratefully acknowledge the contribution of the reviewers' comments.

## 9 REFERENCES

[1] R. A. Calvo and S. D'Mello, "Affect detection: An interdisciplinary review of models, methods, and their applications," *IEEE Transactions on Affective Computing*, vol. 1, no. 1, pp. 18-37, 2010.

[2] C. Izard, "Innate and Universal Facial Expressions: Evidence from Developmental and Cross-Cultural Research," *Psychological Bull.*, vol. 115, pp. 288-299, 1994.

[3] P. Ekman, W. V. Friesen and J. C. Hager, "FACS Manual," Salt Lake City, UT: A Human Face, May 2002.

[4] J. Whitehill, M. S. Bartlett and J. Movellan, "Automatic facial expression recognition," in *Social Emotions in Nature and Artifact*, Oxford University Press, 2013.

[5] Y. Chang, C. Hu and M. Turk, "Manifold of facial expression," *IEEE International Workshop on Analysis and Modeling of Faces and Gestures*, p. 28–35, 2003.

[6] M. Pantic and I. Patras, "Dynamics of facial expression: Recognition of facial ac- tions and their temporal segments from face profile image sequences," *IEEE Transactions on Systems, Man, and Cybernetics*, vol. 36, no. 2, p. 433–449, 2006.

[7] M. Pantic and L. Rothkrantz, "Facial action recognition for facial expression analysis from static face images," *IEEE Transactions on Systems, Man, and Cybernetics*, vol. 34, no. 3, p. 1449–1461, 2004.

[8] I. Cohen, N. Sebe, A. Garg, L. Chen and T. Huang, "Facial expression recognition from video sequences: Temporal and static modeling," *Comput. Vis. Image Understand.*, vol. 91, p. 160–187, 2003.

[9] S. M. Lajevardi and Z. M. Hussain, "Automatic facial expression recognition: feature extraction and selection," *Signal, Image and Video Processing*, vol. 6, no. 1, pp. 159-169, 2012.

[10] G. Zhao and M. Pietikainen, "Dynamic texture recognition using local binary patterns with an application to facial expressions," *IEEE Trans. Pattern Anal. Mach. Intell.*, vol. 29, no. 6, p. 915–928, 2007.

[11] T. Jabid, M. Kabir and O. Chae, "Robust facial expression recognition based on local directional pattern," *ETRI Journal*, vol. 32, pp. 784-794, 2010.

[12] M. H. Kabir, T. Jabid and O. Chae, "A Local Directional Pattern Variance (LDPv) based Face Descriptor for Human Facial Expression Recognition," *7th IEEE Int. Conf. on Advanced Video and Signal Based Surveillance*, pp. 526-532, 2010 .

[13] C. Shan and R. Braspenning, "Recognizing facial expressions automatically from video," in *Handbook of ambient intelligence and smart environments*, 2010, pp. 479-509.

[14] C. Shan, S. Gong and P. W. McOwan, "Robust facial expression recognition using local binary patterns," *IEEE International Conference on Image Processing*, 2005.

[15] C. Shan and T. Gritti, "Learning Discriminative LBP-Histogram Bins for Facial Expression Recognition," *British Machine Vision Conference*, 2008.

[16] M. Song, D. Tao, Z. Liu, X. Li and M. Zhou, "Image ratio features for facial expression recognition application," *IEEE Transactions on Systems, Man, and Cybernetics, Part B: Cybernetics*, vol. 40, no. 3, pp. 779-788, 2010.

[17] L. Zhang and D. Tjondronegoro, "Facial expression recognition using facial movement features," *IEEE Transactions on Affective Computing*, vol. 2, no. 4, pp. 219-229, 2011.

[18] Y. Zhang and Q. Ji, "Active and dynamic information fusion for facial expression understanding from image sequences," *IEEE Transactions on Pattern Analysis and Machine Intelligence*, vol. 27, no. 5, pp. 699-714, 2005.

[19] Y. Tian, T. Kanade and J. F. Cohn, "Recognizing lower face action units for facial expression analysis," *IEEE International Conference on Automatic Face and Gesture Recognition*, pp. 484-490, 2000.

[20] M. Z. Uddin, J. J. Lee and T.-S. Kim, "An enhanced independent component-based human facial expression recognition from video," *IEEE Transactions on Consumer Electronics*, vol. 55, no. 4, pp. 2216-2224, 2009.

[21] M. F. Valstar and M. Pantic, "Combined support vector machines and hidden markov models for modeling facial action temporal dynamics," *Human–Computer Interaction*, pp. 118-127,

2007.

[22] T. Cootes, G. Edwards and C. Taylor, "Active appearance models," *IEEE Trans. on Pattern Analysis and Machine Intelligence,* vol. 23, no. 6, pp. 681--685, 2001.

[23] S. Lucey, I. Matthews, C. Hu, Z. Ambadar, F. D. l. Torre and J. Cohn, "AAM derived face representations for robust facial action recognition," *7th International Conference on Automatic Face and Gesture Recognition,* 2006.

[24] A. B. Ashraf, S. Lucey, T. C. Jeffrey F. Cohn, Z. Ambadar, K. M. Prkachin and P. E. Solomon, "The painful face–pain expression recognition using active appearance models," *Image and Vision Computing,* vol. 27, no. 12, pp. 1788-1796, 2009.

[25] B. Abboud, F. Davoine and M. Dang, "Facial expression recognition and synthesis based on an appearance model," *Signal Processing: Image Communication,* vol. 19, no. 8, pp. 723-740, 2004.

[26] A. Asthana, J. Saragih, M. Wagner and R. Goecke, "Evaluating aam fitting methods for facial expression recognition," *3rd International Conference on Affective Computing and Intelligent Interaction and Workshops,* pp. 1-8, 2009.

[27] C. Martin, U. Werner and H.-M. Gross, "A real-time facial expression recognition system based on active appearance models using gray images and edge images," *8th IEEE International Conference on Automatic Face & Gesture Recognition,* 2008.

[28] D. Cristinacce and T. F. Cootes, "Feature Detection and Tracking with Constrained Local Models," in *British Machine Vision Conference,* 2006.

[29] J. M. Saragih, S. Lucey and J. F. Cohn., "Deformable model fitting by regularized landmark mean-shift," *International Journal of Computer Vision,* vol. 91, no. 2, pp. 200-215, 2011.

[30] A. Asthana, S. Zafeiriou, S. Cheng and M. Pantic, "Robust discriminative response map fitting with constrained local models," in *IEEE Conference on Computer Vision and Pattern Recognition,* 2013.

[31] S. W. Chew, P. Lucey, S. Lucey, J. Saragih, J. F. Cohn, I. Matthews and S. Sridharan, "In the pursuit of effective affective computing: The relationship between features and registration," *IEEE Transactions on Systems, Man, and Cybernetics, Part B: Cybernetics,* vol. 42, no. 4, pp. 1006-1016, 2012.

[32] M. Bartlett, G. Littlewort, M. Frank, C. Lainscsek, I. Fasel and J. Movellan, "Recognizing facial expression: machine learning and application to spontaneous behavior," in *IEEE Computer Society Conf. on Computer Vision and Pattern Recognition,* 2005.

[33] M. Lyons, J. Budynek and S. Akamatsu, "Automatic classification of single facial images," *IEEE Trans. on Pattern Analysis and Machine Intelligence,* vol. 21, no. 12, pp. 1357 - 1362 , Dec 1999.

[34] A. Hadid, M. Pietikainen and T. Ahonen, "A discriminative feature space for detecting and recognizing faces," in *IEEE Computer Society Conf. on Computer Vision and Pattern Recognition,* 2004.

[35] S. L. Happy, A. George and A. Routray, "A real time facial expression classification system using Local Binary Patterns," *4th Int. Conf. on Intelligent Human Computer Interaction,* 2012.

[36] A. Dhall, A. Asthana, R. Goecke and T. Gedeon, "Emotion recognition using PHOG and LPQ features," in *IEEE International Conference on Automatic Face and Gesture Recognition and Workshops ,* 2011.

[37] C. Shan, S. Gong and P. W. McOwan, "Facial expression recognition based on local binary patterns: A comprehensive study," *Image and Vision Computing,* vol. 27, no. 6, pp. 803-816, 2009.

[38] K. I. Kim, K. Jung and H. J. Kim, "Face recognition using kernel principal component analysis," *IEEE Signal Processing Letters,* vol. 9, no. 2, pp. 40-42, 2002.

[39] A. J. Calder, A. M. Burton, P. Miller, A. W. Young and S. Akamatsu, "A principal component analysis of facial expressions," *Vision research,* vol. 41, no. 9, pp. 1179-1208, 2001.

[40] H.-B. Deng, L.-W. Jin, L.-X. Zhen and J.-C. Huang, "A new facial expression recognition method based on local gabor filter bank and pca plus lda," *International Journal of Information Technology,* vol. 11, no. 11, pp. 86-96, 2005.

[41] Z. Zhang, Y. Yan and H. Wang, " Discriminative filter based regression learning for facial expression recognition," *20th IEEE International Conference on Image Processing (ICIP),* pp. 1192 - 1196 , 2013 .

[42] C. Shan, S. Gong and P. W. McOwan, "A comprehensive empirical study on linear subspace methods for facial expression analysis," *IEEE Conf. on Computer Vision and Pattern Recognition Workshop,* pp. 153-153, 2006.

[43] S.-K. Oh, S.-H. Yoo and W. Pedrycz, "Design of face recognition algorithm using PCA-LDA combined for hybrid data pre-processing and polynomial-based RBF neural networks: Design and its application," *Expert Systems with Applications,* vol. 40, no. 5, pp. 1451-1466, 2013.

[44] Y. Rahulamathavan, R.-W. P. J. A. Chambers and D. J. Parish, "Facial Expression Recognition in the Encrypted Domain Based on Local Fisher Discriminant Analysis," *IEEE Transactions on Affective Computing,* vol. 4, no. 1, pp. 83-92, 2013.

[45] H. Mohammadzade and D. Hatzinakos, "Projection into expression subspaces for face recognition from single sample per person," *IEEE Transactions on Affective Computing,* vol. 4, no. 1, pp. 69-82, 2013.

[46] S. Moore and R. Bowden, "Local binary patterns for multi-view facial expression recognition," *Computer Vision and Image Understanding,* vol. 115, no. 4, pp. 541-558, 2011.

[47] L. Zhong, Q. Liu, P. Yang, B. Liu, J. Huang and D. N. Metaxas, "Learning active facial patches for expression analysis," in *IEEE Conference on Computer Vision and Pattern Recognition (CVPR),* 2012.

[48] L. Unzueta, W. Pimenta, J. Goenetxea, L. P. Santos and F. Dornaika, "Efficient generic face model fitting to images and videos," *Image and Vision Computing,* vol. 32, no. 5, pp. 321-334, 2014.

[49] P. Viola and M. Jones, "Rapid object detection using a boosted cascade of simple features," in *IEEE Conference on Computer Vision and Pattern Recognition,* 2001.

[50] D. Nguyen, D. Halupka, P. Aarabi and A. Sheikholeslami, "Real-time face detection and lip feature extraction using field-programmable gate arrays," *IEEE Transactions on Systems, Man, and Cybernetics, Part B: Cybernetics,* vol. 36, no. 4, pp. 902-912,

2006.

[51] R. C. Gonzalez and R. E. Woods, Digital Image Processing (3rd Edition), Pearson Education, 2008.

[52] N. Otsu, "A threshold selection method from gray-level histograms," *IEEE Transactions on Systems, Man and Cybernetics,* vol. 9, no. 1, pp. 62-66, 1979.

[53] T. Ojala, M. Pietikainen and T. Maenpaa, "Multiresolution gray-scale and rotation invariant texture classification with local binary patterns," *IEEE Transactions on Pattern Analysis and Machine Intelligence,* vol. 24, no. 7, pp. 971-987, 2002.

[54] P. N. Belhumeur, J. P. Hespanha and D. Kriegman, "Eigenfaces vs. fisherfaces: Recognition using class specific linear projection," *IEEE Transactions on Pattern Analysis and Machine Intelligence,* vol. 19, no. 7, pp. 711-720, 1997.

[55] C. Cortes and V. Vapnik, "Support-vector networks," *Machine learning,* vol. 20, no. 3, pp. 273-297, 1995.

[56] C.-W. Hsu and C.-J. Lin, "A comparison of methods for multiclass support vector machines," *IEEE Transactions on Neural Networks,* vol. 13, no. 2, pp. 415-425, 2002.

[57] M. J. Lyons, M. Kamachi and J. Gyoba, "Japanese Female Facial Expressions (JAFFE)," Database of digital images, 1997.

[58] P. Lucey, J. F. Cohn, T. Kanade, J. Saragih, Z. Ambadar and I. Matthews, "The Extended Cohn-Kande Dataset (CK+): A complete facial expression dataset for action unit and emotion-specified expression," in *3rd IEEE Workshop on CVPR for Human Communicative Behavior Analysis,* 2010.

[59] Y. Tian, T. Kanade and J. F. Cohn, "Facial expression recognition," in *Handbook of face recognition,* London, Springer , 2011, pp. 487-519.

[60] M. Sokolova and G. Lapalme, "A systematic analysis of performance measures for classification tasks," *Information Processing & Management,* vol. 45, no. 4, pp. 427-437, 2009.

[61] P. A, N. H.A., S. M.G. and Y. N., "Gauss–Laguerre wavelet textural feature fusion with geometrical information for facial expression identification," *EURASIP Journal on Image and Video Processing,* 2012.

[62] Z. Zhang, C. Fang and X. Ding, "Facial expression analysis across databases," *IEEE International Conference on Multimedia Technology,* pp. 317-320, 2011.

[63] O. Jesorsky, K. Kirchberg and R. Frischholz, "Robust face detection using the hausdorff distance," in *3rd International Conference on Audio- and Video-Based Biometric Person Authentication,* Halmstad, Sweden, 2001.

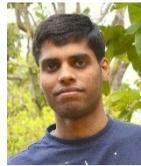

**S L Happy** has received the B.Tech. (Hons.) degree from Institute of Technical Education and Research (ITER), India in 2011. Now he is pursuing the M. S. degree from Indian Institute of Technology Kharagpur, India. His research interests include pattern recognition, computer vision and facial expression analysis.

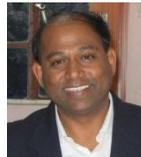

**Aurobinda Routray** has received his Masters degrees in 1991 from IIT Kanpur, India and his PhD in 1999 from Sambalpur University, India. He has also worked as a postdoctoral researcher at Purdue University, USA, during 2003-2004. He is currently working as a professor in the Department of Electrical Engineering, Indian Institute of Technology, Kharagpur. His research interests include non-linear and statistical signal processing, signal based fault detection and diagnosis, real time and embedded signal processing, numerical linear algebra, and data driven diagnostics.